\documentclass[10pt,twocolumn,letterpaper]{article}

\usepackage{cvpr}
\usepackage{times}
\usepackage{epsfig}
\usepackage{graphicx}
\usepackage{amsmath}
\usepackage{amssymb}
\usepackage{adjustbox}

\usepackage[pagebackref=true,breaklinks=true,letterpaper=true,colorlinks,bookmarks=false]{hyperref}

\cvprfinalcopy 

\def\cvprPaperID{****} 

\ifcvprfinal\fi
\begin{document}

\title{SHAQ: Single Headed Attention with Quasi-Recurrence}

\author{Nashwin Bharwani\\
{\tt\small nbharwani3@gatech.org}
\and
Warren Kushner\\
{\tt\small wkushner3@gatech.edu}
\and
Sangeet Dandona\\
{\tt\small sdandona3@gatech.edu}
\and
Ben Schreiber\\
{\tt\small bschreiber3@gatech.edu}
}

\maketitle


\begin{abstract}
    Natural Language Processing research has recently been dominated by large scale transformer models. Although they achieve state of the art on many important language tasks, transformers often require expensive compute resources, and days spanning to weeks to train. This is feasible for researchers at big tech companies and leading research universities, but not for scrappy start-up founders, students, and independent researchers. Stephen Merity's SHA-RNN, a compact, hybrid attention-RNN model, is designed for consumer-grade modeling as it requires significantly fewer parameters and less training time to reach near state of the art results. We analyze Merity's model here through an exploratory model analysis over several units of the architecture considering both training time and overall quality in our assessment. Ultimately, we combine these findings into a new architecture which we call SHAQ: Single Headed Attention Quasi-recurrent Neural Network. With our new architecture we achieved similar accuracy results as the SHA-RNN while accomplishing a 4x speed boost in training. 
\end{abstract}

\section{Introduction}

This paper presents a model analysis of the Single Headed Attention Recurrent Neural Network (SHA-RNN) architecture popularized by Stephen Merity in his paper \textit{Single Headed Attention RNN: Stop Thinking With Your Head}~\cite{merity2019single}. In this paper Merity reports near state of the art performance on the ENWIK8 character modeling task using a single GPU (Titan V) inside of one day. 

When Sha-RNN was published the current state of the art (SOTA) in language modeling was becoming increasingly led by transformer models. Merity's work demonstrated that it was possible to achieve near-SOTA results without relying on large, complex transformer architectures. Our aim here is to extend Merity's research of the SHA-RNN by seeing if we can enhance the original network to increase training speed and/or accuracy while minimizing the memory footprint.

Although Merity's results were widely celebrated in the practitioner community, the paper itself has limited detail on how the author actually arrived at the final architecture. This is not a problem for those who merely want to use the model. However, for those who want to understand and further optimize the model, it is important to know what each component's purpose is and how it impacts performance. To address this need, we performed an ablation study over several components of the model, both by removing them and substituting in new components. More specifically, we wanted to examine the recurrent layer of the network, boom layer, custom attention head layer, and number of layer blocks in the network. 
By applying SHA-RNN to the ENWIK8 dataset we measure both the quality of solutions that our ablation studies produce and the amount of training time it takes to achieve them. Additionally, from the results of our experimentation, we introduce a new variation on SHA-RNN called the Single Headed Attention Quasi-recurrent Neural Network (SHAQ) which has shown to outperform the SHA-RNN model in terms of both speed and accuracy.

\subsection{Background and Motivation}
The application of interest here is character prediction which falls under a more broad category of Natural Language Processing (NLP), in which transformer networks have dominated. These networks are able to make use of ordered information through positional embeddings without the need to run sequentially. This allows for parallelization in both training and inference time. Additionally, transformers develop a strong sense of context awareness through their attention mechanism, which consists of estimating a distribution of the importance of preceding tokens to inform the prediction of the next token~\cite{vaswani2017attention}.

Transformers have achieved state of the art results in many problems including machine translation, text prediction and conversation. One of the most notable transformer architectures is BERT, which is a bi-directional encoder model which means that it uses both left oriented and right oriented context to predict tokens in a sentence~\cite{devlin2019bert}. It has achieved record-breaking performance on important language problems including GLUE, a composite score of an array of natural language tasks, and SQuAD, which involves predicting the answer to a question given a Wikipedia passage containing necessary information. 

Since BERT was released, it has become a common pretrained foundation for many practical language applications and has served as a baseline and fountain of inspiration for other cutting edge research projects. One example of this is RoBerta, where the team started with the BERT model architecture and made several adjustments during training including training on a 10x larger input data set, increasing the batch size, increasing the vocabulary size, and dynamically changing the masks applied to the training examples~\cite{liu2019roberta}. This achieved an even better GLUE score than the original BERT, and outperformed it on each of the 9 tasks.

One important thing to note about models like BERT and RoBERTa is that they are driven by abundantly funded organizations and institutions such as Google, Facebook, and University of Washington. While there is merit in striving for the absolute maximum performance that can be achieved with massive scale transformers, it is often infeasible for startups and individuals on a bootstrap budget to build on or even apply this research, limiting the audience to which it is relevant. These limitations bottleneck innovation, preventing many researchers from building upon previous advancements and limiting the influx of ideas. In particular, outside the realm of research and academia there are constraints of budget, training time, and machine resources which make it difficult to engineer something like BERT which requires 16 TPU's and 4 days to train, and resulted in a 340M parameter model, which will not be trivial to store and serve at-scale in production. If successful, this paper will show how SHA-RNN can be customized for use by researchers with limited computing resources. The introduction of SHAQ reinforces Merity's argument that that you do not need to build extremely complex models to reach near SOTA results.

Stephen Merity notes that there is a significant bias in Big Tech and academia towards enormous transformer models remarking how researchers are trending towards this approach like "moths to a light-bulb"~\cite{merity2019single}. On top of this, Merity also reminds us that there has not been a definitive answer as to whether RNN models are inferior to transformers especially in low-resource situations. It turns out that the door really was not shut for RNN's. Using a multi-layer RNN architecture that employs an attention head mechanism, Merity achieves excellent results on a single GPU with one day of training and a 50M parameter model.

Our goal in this endeavor is to understand and build on what Merity has started with SHA-RNN. Our ablation study shows how some components are critical to success while others can be swapped out. Furthermore, SHA-RNN can be scaled in such a way that it becomes much leaner and faster to train at the cost of slight performance degradation. This kind of information is useful for people who are tight on resources so that they can make the best decision possible when it comes to training, serving, and storing their models.
\subsection{ENWIK8 Dataset}
We employed the ENWIK8 dataset, which is the main focus of the Hutter Prize competition.  Regarding the aspects mentioned in \cite{gebru2020datasheets}, we will consider motivation, composition, collection and preprocessing. The motivation of this dataset by computer scientist Marcus Hutter was to ignite AI innovation through an equivalent task of compression. He chose to use Wikipedia as a corpus since it is a good proxy for a snapshot of human knowledge at a given moment in time. The ENWIK8 dataset is 100M UTF encoded bytes that were from an XML dump of the English version of Wikipedia. 

In terms of preprocessing, we were able to take full advantage of the utilities in the Merity code base which splits train, validation and test sets of 90M, 5M, and 5M respectively, and cuts the data into variable length character sequences of a minimum of 5 characters long. The user has some control over these sequence lengths by providing a center of a distribution in which the lengths are drawn at random.  With respect to collection, this dataset will have any of the same flaws that are present in Wikipedia pages concerning sensitivity and privacy.

\section{Architecture Overview}

\subsection{SHA-RNN}

The SHA-RNN architecture was inspired by transformers and adopts an attention head and fully connected layers to support a recurrent cell. The network itself is composed of an embedding layer followed by four SHA-RNN blocks. Each block contains an LSTM cell, followed by a custom attention head and a modification of a fully-connected layer which Merity refers to as a 'Boom layer'~\cite{merity2019single}. Both the attention head and the Boom layer have a residual connection.

\begin{figure}
\includegraphics[scale=0.35]{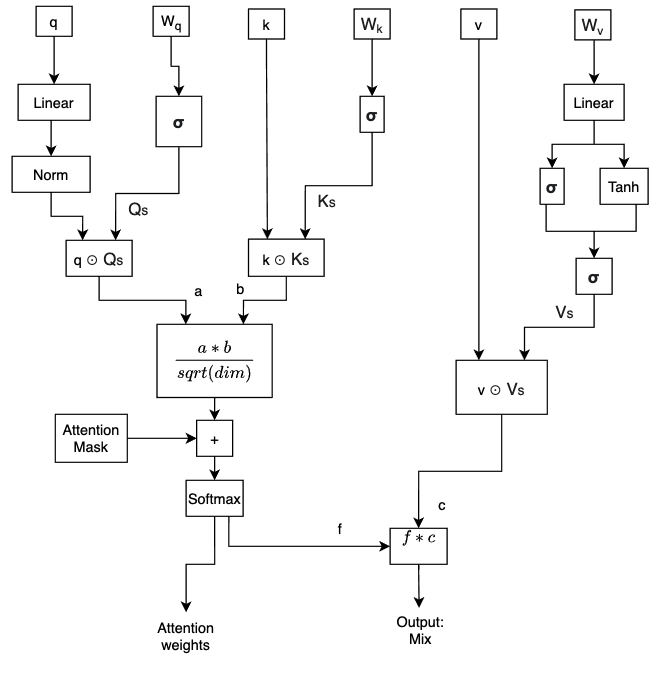}
\caption{S. Merity's custom attention head. This differs from traditional attention since it does not apply a linear layer to the $k$ and $v$ inputs, thereby reducing the number of parameters to train.}
    \label{fig:attn}
\end{figure}

The custom attention head primarily differs from standard attention heads by ignoring the linear pre-processing step when preparing the key and value parameters. It incorporates 3 gating vectors called $qs$, $ks$  and $vs$ which are multiplied elementwise across the entire set of the corresponding query, key and value vectors. It also adds Dropout and Normalization layers in several sections to help regularize the network. This attention head can be approximated visually by figure \ref{fig:attn}.

The Boom layer is a simplification of a typical feed-forward fully-connected cell commonly used in transformer architectures with linear layers which expands dimensions 1024 to 2048, a GeLU activation function, and a linear layer that goes from 2048 to 1024. Merity's approach replaces the second layer with an operation to split the activation function output into multiple chunks, then sum them together.
This has empirically shown to be a good approximation of the linear layer by producing slightly worse results, by significantly reducing the number of parameters and increasing speed, leading to a more efficient architecture.

The SHA-RNN model is trained using the LAMB optimizer. This optimizer often works significantly better than standard SGD optimizers on transformer models, and in practice has been able to reduce the training time of BERT from 3 days to 76 minutes.~\cite{you2020large} It accomplishes this as a modification of the Layerwise Adaptive Rate Scaling (LARS) optimizer, making a few modifications to the trust ratio, and adopting the ADAM update rule.~\cite{mann_2019}

This optimizes using the Cross Entropy Loss after a softmax layer. The network performance is measured using bits-per-character (bpc), which is the average number of bits necessary to represent a character~\cite{huyen_2019}.  A smaller number bits means a more efficient encoding~\cite{huyen_2019}.

\begin{figure}
    \centering
    \includegraphics[scale=0.5]{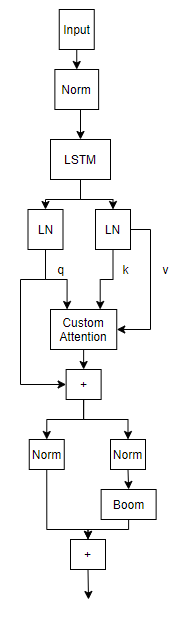}
    \caption{A single Sha-Rnn block. Multiple blocks are chained together in sequence to train the full network. In addition to the LSTM, Merity includes his custom attention head and Boom layer in order to process the output of the recurrent cell}
    \label{fig:sha-rnn}
\end{figure}

\section{Approach}




We performed a series of experiments based on Merity's original implementation of the SHA-RNN~\cite{smerity}. We first ran a baseline of Merity's original code to capture bpc and time per epoch metrics and reported this as our baseline for comparison. Since we used Merity's exact implementation we reasoned any deviation from the reported results in the original paper were due to slight differences in training and hyperparameter tweaks and not mistakes in our implementation of the architecture. In the paper, Merity implemented multiple novel modifications both to individual layers (like self-attention) as well as adding his own unique layer he called Boom. He received very impressive scores, but it was not clear how much of an influence each innovation he made had. He also did not report an ablation study in many aspects of the architecture. In order to decipher which layers and modifications had impact, we performed our own ablation study. 

In addition, we wanted to see if we could then improve upon the original architecture. Using the results of the ablation study along with testing hypotheses inspired by the literature for optimizing sequential networks, we made further custom additions that we thought could yield positive results on bpc and time efficiency. We then compared our modifications with the baseline findings to see if our additions were fruitful. 

Finally, we included our modification in the ablation/modification studies that either decreased model parameters, time efficiency and/or bpc score for a final architecture and recorded the results.


In designing these experiments, we believed they would reveal interesting properties of SHA-RNN because ablation studies are prevalent and proven methods to analyze existing deep learning architectures. We also felt we could improve SHA-RNN because, as Merity notes, the SHA-RNN architecture was put together rather quickly without a complete analysis of tradeoffs. This suggested that there was an opportunity for optimization and refinement of the existing model. In SHAQ, our proposed architecture, we made use of the convolutional nature of Quasi-recurrent Neural Networks (QRNN) in addition to other optimizations to provide a faster and more robust model.

\section{Experiment and Results}

Specific experiments were chosen in order to investigate the purpose and performance of the individual components constructing the SHA-RNN architecture. As such, experiments were divided into Boom analysis, Attention analysis, Layer analysis, and Recurrent Unit analysis sections.

All experiments were run for 32 epochs and were compared against a baseline, Merity's SHA-RNN model using the default parameters published to his Github repository~\cite{smerity}. For consistency, all comparisons were performed with the baselines originating from the same GPU type. An experiment was considered successful if it is able to outperform the baseline in either time or performance without degradation. This was explicitly defined as offering a better bpc on the test set given a similar amount of training time per epoch or offering a similar bpc on the test set and a smaller training time per epoch. Experiments were performed using Merity's Pytorch implementation on his Github repository~\cite{smerity} as a starting point and making minor modifications to it. These updates can be found on our code repository \footnote{https://github.gatech.edu/deep-learning-gatech}. 

\begin{table}[]
    \centering
    \resizebox{\columnwidth}{!}{%
    \begin{tabular}{|c|c|c|c|c|}
        \hline \textbf{Experiment} & \textbf{avg. Time/Epoch} & \textbf{Params} & \textbf{Loss} & \textbf{bpc} \\
        \hline Baseline & 2.16h & 54M & 0.84 & 1.208 \\
        \hline Removed Boom & 1.84h & 37M & 0.78 & 1.120 \\
        \hline Replace Boom with FC & 2.33h & 71M & 0.82 & 1.158 \\
        \hline QRNN (w=2) & 0.92h & 45M & 0.78 & 1.126 \\
        \hline Mean Attention & 2.03h & 53M & 0.84  & 1.208 \\
        \hline Removal of $Qs$,$Ks$,$Vs$ & 2.12h & 51M & 0.78  & 1.13 \\
        \hline
    \end{tabular}
    }
    \caption{Experiment results performed on an Tesla v100. All experiments were conducted with the same parameters and varying elements of the architecture. The accuracy metrics (.ie loss and bpc) were from the test set after 32 epochs.}
    \label{tab:exp_results}
\end{table}

\subsection{Boom}

In order to determine the effectiveness and contribution of the Boom layer, we performed two experiments. The first of which replaced the Boom layer with its analogous Fully Connected layer and the latter of which removed the Boom layer entirely. We hypothesized that the fully-connected layer would eventually converge to a better value after all 32 epochs, but the Boom layer would outperform it after the same amount of time elapsed. Additionally, we believed that having no boom layer would cause the network to converge faster, but with a worse bpc. 

We found that the Baseline model using Merity's Boom layer was able to outperform the fully connected model until approximately the 26 hour mark. This showed that removing complexity from the model allowed the network to optimize more effectively when the model is trained under significant time pressure. It also demonstrated that if training time was not a factor, the Fully Connected layer would perform better.

Surprisingly though, the network performed better and trained faster than the baseline without including any fully connected components to process the results of the attention head. This proved to be a successful experiment as it surpassed our expectations and demonstrated that adding a component after the attention head actively made the network worse. This is likely because the network is not generalizing well enough to compensate for the additional complexity.

\begin{figure}
    \centering
    \includegraphics[scale=0.6]{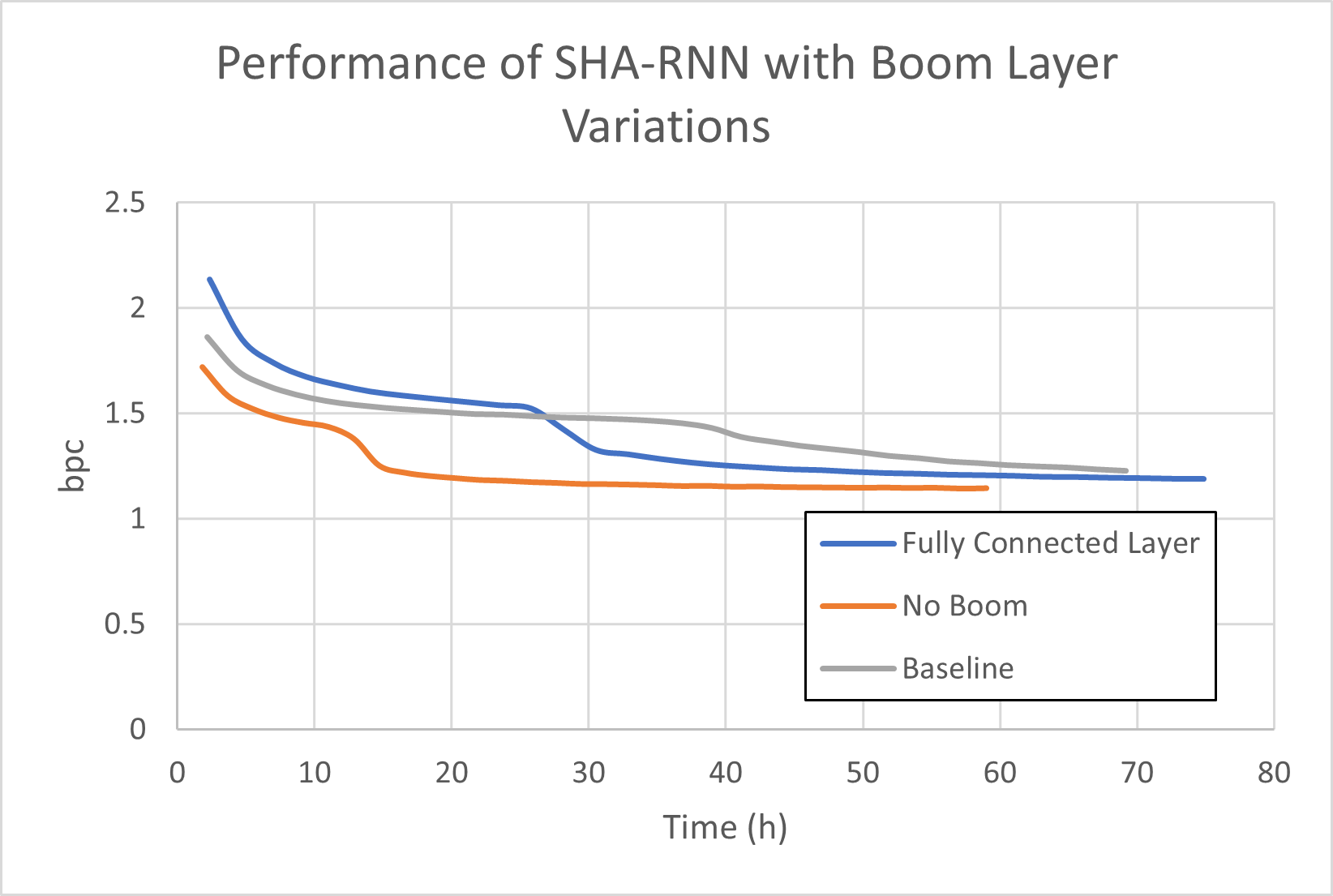}
    \caption{Experiments to determine the value of a Boom layer on a Tesla V100.}
    \label{fig:boom_graph}
\end{figure}

\subsection{Alternate-Attention Head Experiments }

\begin{figure}
    \centering
    \includegraphics[scale=0.35]{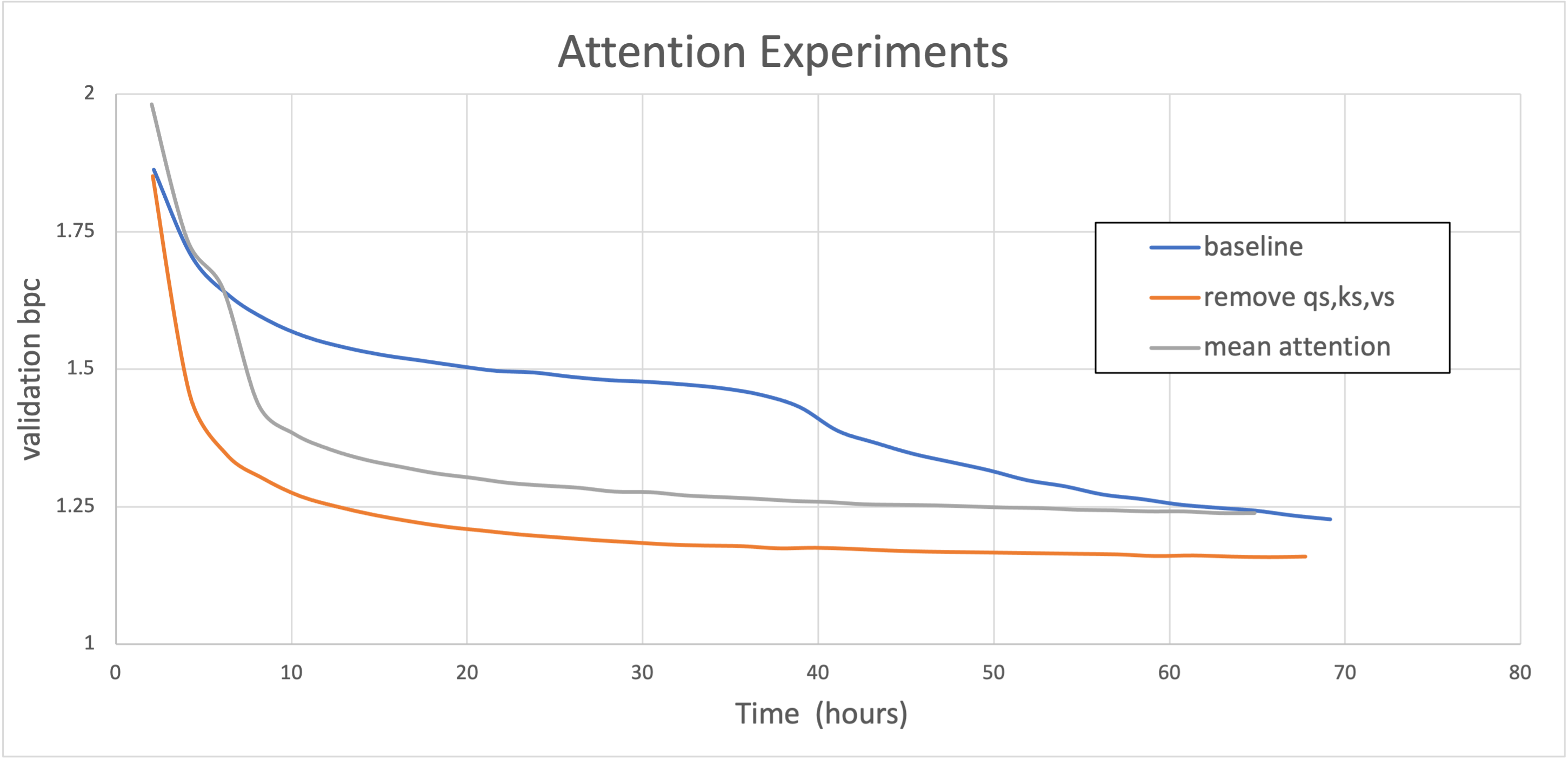}
    \caption{Displays Experiments altering Merity's attention layer and charting the validation bpc results over training time of 32 epochs in hours on a Tesla V100.}
    \label{fig:layerattn_merity}
\end{figure}

Merity used a custom self-attention head illustrated in Figure~\ref{fig:attn}.
In his implementation every query, key and value had an accompanied $Q_{s}$, $K_{s}$ and $V_{s}$ vector. These were created by having their own trainable parameters with a sigmoid activation to bind these ranges from [0, 1]. Figure~\ref{fig:attn} shows that they are multiplied elementwise to their corresponding Q, K, and V vector counterparts. Merity reasoned that certain dimensions in the LSTM hidden vectors contain local information while others encapsulate global information. As the network trains these parameters, $Q_{s}$, $K_{s}$ and $V_{s}$ decide what dimensions to exclude versus what dimensions to incorporate ~\cite{merity2019single}. This idea, however, was never tested in the original paper. 

For the first attention experiment, $Q_{s}$, $K_{s}$ and $V_{s}$ were removed making the attention layer much simpler. This resulted in a smoother training and a much steeper learning curve (implying faster convergence) than baseline as can be seen in Figure~\ref{fig:layerattn_merity}. Along with the faster convergence, it went through 32 epochs of training 1.5 hours faster than baseline on a Tesla V100. This was due to fewer operations in the attention layer. Finally, it had a significantly better bpc score than the baseline on the test set at 1.131 bpc.

Merity also made use of caching the memory state vectors originating from the LSTM in prior windows of the training loop. For example, if the sequence length of the LSTM is $p$ the LSTM will emit $[X_{i},...,X_{i+p}]$ along with $[M_{i},...,M_{i+p}]$ cell states where each $X$ is the output vector, each $M$ is the memory vector and subscript $i$ is the index of the token within the document. In the training loop all the $M$'s are cached and used for the next forward pass. Since the training loop only increments 1 token at a time the memory is stored and becomes $M'$, where $M'$ is composed of vectors $[M'_{i-1},...,M'_{i+p-1}]$ in relation to position $i$ (the current LSTM position). This cached sequence is concatenated with the current memories produced and the cached memory tensor gets larger until it reaches a max sequence length of $5000-p$. In the attention layer these memories are also concatenated with the current hidden state $H$ by the sequence dimensions to make the key and value vectors \cite{smerity}. As a result of the larger sequence size, the dot product in self attention will have many more operations. The modified mean attention condenses all of these concatenated memory vectors to its mean along the sequence dimension, thus reducing the computation for the downstream dot product and can be seen as follows: 
   \begin{align}
        M^{q} =  [M_{i-q}^{q},...,M_{i+p-q}^{q}], \nonumber \\
        ..., \nonumber \\
        M'' =  [M''_{i-2},...,M''_{i+p-2}], \nonumber \\
        M' =  [M'_{i-1},...,M'_{i+p-1}], \nonumber \\
        C = Concat([M^{q},...,M'',M'])
    \end{align}
    \begin{equation}
        W = 1/(p*q) \sum\limits_{j=0}^{p*q}   C_{j}
    \end{equation}
    \begin{equation}
       Z = [W, H_{i},...,H_{i+p} ]
    \end{equation}
    \begin{equation}
         key, val \approx  Z, Z 
     \end{equation}
Where all $H$, $M$, and $W$ vectors have a size of 1024 ($\approx$ denotes downstream operations for key/val). The above idea was inspired by another simplified attention implementation where they replaced the fully connected operations for Q, K and V with an elementwise multiplication with trainable parameters summed over time \cite{luo2020simplified}. As can be seen in Figure~\ref{fig:layerattn_merity} this did have much smoother convergence compared to baseline. It also completed training time in 64 hours which is a 4 hour speed boost compared to baseline. However, it suffered from a nontrivially worse score from baseline with a 1.219 bpc on the test set.

\subsection{Layer-Attention Analysis}
\begin{figure}
    \centering
    \includegraphics[scale=0.6]{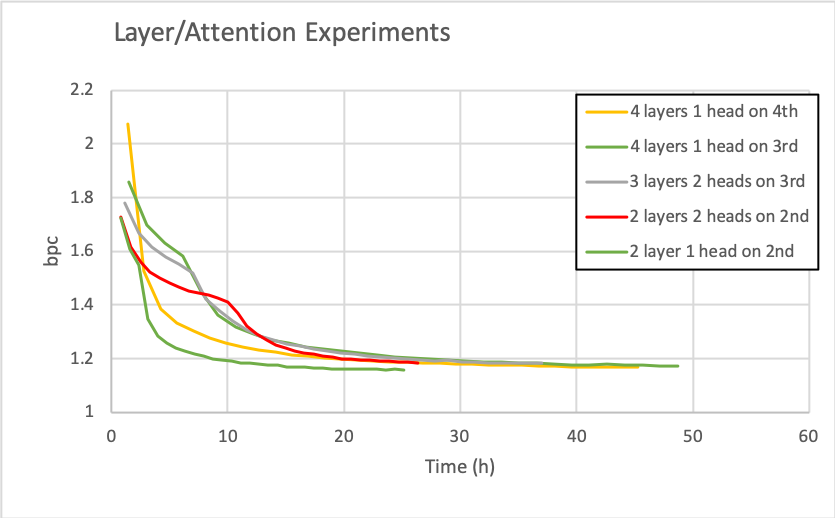}
    \caption{Experiments to understand layer scaling on an RTX 3090.}
    \label{fig:layerattn}
\end{figure}
This line of experiments began by considering why Merity chose to put an attention head specifically on the third layer of the network. We considered that putting it on a different layer could provide some benefit and tested this hypothesis by measuring performance after placing a single attention head on each layer each in a separate experiment. In table \ref{tab:head_layer} are the results, where putting the head on the first two layers performed very poorly and attention on layer 3 and 4 performed relatively the same. As to why this happened, we posit that the attention signal was probably being lost through the earlier layers and the gradients were too small to provide any meaningful update. This was not overfitting, it just was not learning.

Since putting the attention head on either of the first two layers was  unsuccessful, we wondered if these layers were worth the time and memory they consume. We proposed removing one layer and adding a second attention head in the fourth layer would be a reasonable tradeoff. This is since the extra attention head could compensate for the inferential power that we would lose by removing the layer, as more attention heads can jointly attend to information in different representation subspaces \cite{vaswani2017attention}. The results here showed a trade-off in favor of removing the layer in that model quality was stable, but training time was greatly reduced. We then decided to take off another layer without even adding another attention head, which resulted in another speed up without much bpc increase. 

Lastly, we wondered how much worse the 2 layer model would do without its second head. To our surprise, this model performed better than even the baseline model, and achieved these results in about 40\% less time (see figure \ref{fig:layerattn}). This may have been a case of Occam's Razor, where we found the right balance between model complexity and efficiency. However, given that in other research endeavors on ENWIK8, giant transformer models have done well \cite{alrfou2018characterlevel}, it could also be a case of not fine-tuning the more complex model enough. Overall, we deem these experiments a success because we discovered a model architecture which could outperform the baseline in bpc, and require less training time and fewer parameters.

\subsection{QRNN}

\begin{figure}
    \centering
    \includegraphics[scale=0.4]{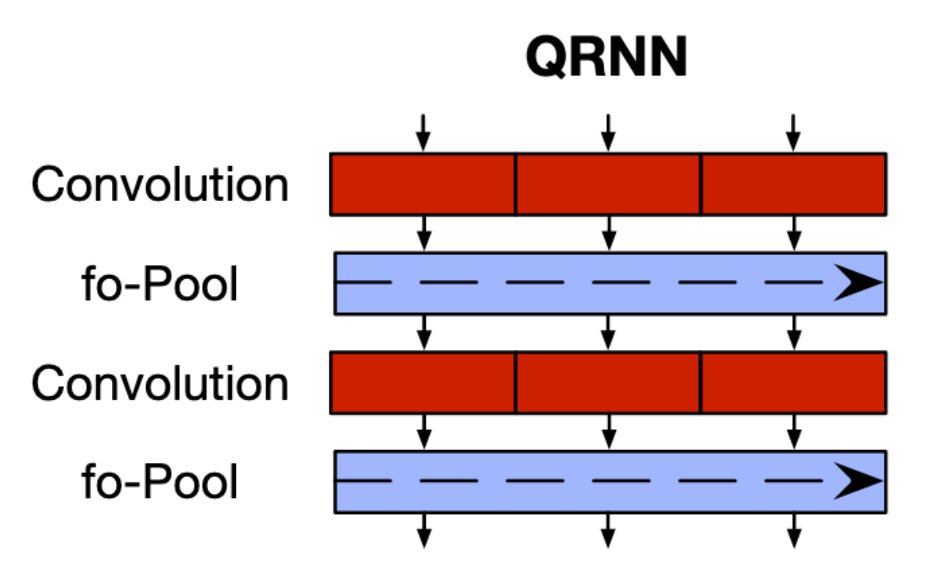}
    \caption{Convolution feeding into fo-pooling~\cite{bradbury2016quasirecurrent}}
    \label{fig:QRNN}
\end{figure}

The original implementation of Merity's code made use of LSTMs as the recurrent unit in each block. We wanted to explore what would happen if the recurrent unit was altered and see if there would be any improvement in the overall speed or performance of the model on ENWIK8. Upon researching into different types of recurrent units, we came across the quasi-recurrent neural network (QRNN) proposed by Bradbary et al.~\cite{bradbury2016quasirecurrent}. The QRNN works by making use of two layers, a convolutional layer and a quasi-recurrent pooling layer as shown in figure \ref{fig:QRNN}.

The basic idea is that the convolutional layer will perform a causal 1d convolution over token embeddings of a sequence. The causal 1d convolution is simply providing a padding layer before the start of the sequence in order to make sure that the network only depends on the previous embeddings or the current one~\cite{youtube1}. Running filters (or kernels) over these embeddings will produce results that emphasize what aspects of the embeddings possess the most information while only looking at the past~\cite{youtube1}. Afterwards, the results of the convolutions are fed into a pooling a layer that preserves the ordering of the tokens. The following equations summarize how the internals of a QRNN works mathematically:

\begin{equation}
    Z = tanh(W_z * X)
    \label{Z}
\end{equation}
\begin{equation}
    F = \sigma(W_f * X)
    \label{F}
\end{equation}
\begin{equation}
    O = \sigma(W_o * X)
    \label{O}
\end{equation}
\begin{equation}
    c_t = f_t \odot c_t-1 + (1-f_t) \odot z_t
    \label{ct}
\end{equation}
\begin{equation}
    h_t = o_t \odot c_t
    \label{ht}
\end{equation}

Equations (\ref{Z}), (\ref{F}), and (\ref{O}) represent the convolutional operations of the cell.  Three different filter types are convolved over the sequence and fed into activations. $Z$, $F$, and $O$ are different gates of the cell representing an input gate, forget gate, and output gate respectively~\cite{jagtap_2020}. Equations (\ref{ct}) and (\ref{ht}) represent the quasi-recurrent pooling layer which preserves ordering and calculates the cell states and hidden states respectively.

\begin{figure}
    \centering
    \includegraphics[scale=0.6]{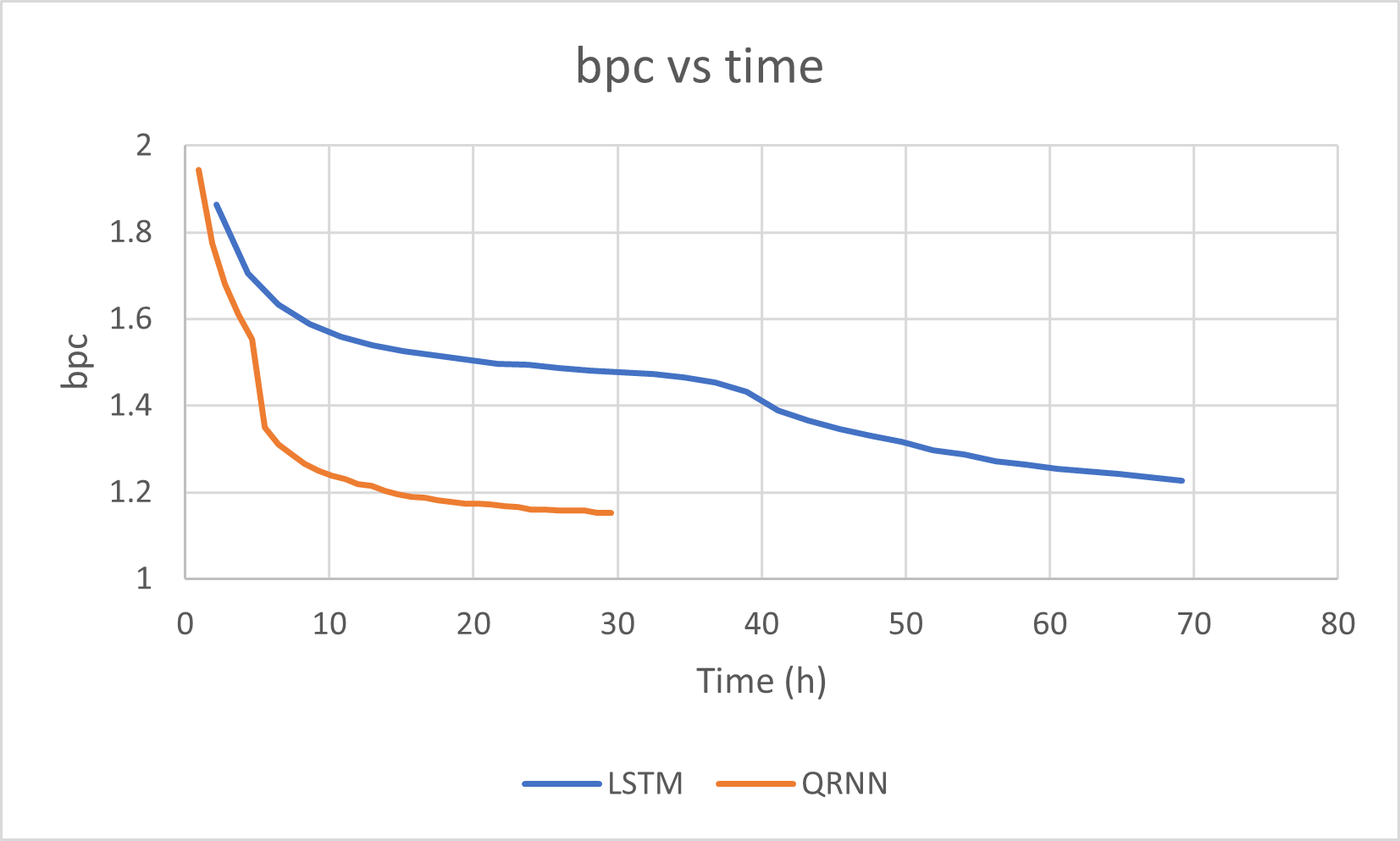}
    \caption{LSTM/QRNN Validation bpc vs time on a Tesla V100.}
    \label{fig:lq_bpc_time}
\end{figure}

\begin{figure}
    \centering
    \includegraphics[scale=0.6]{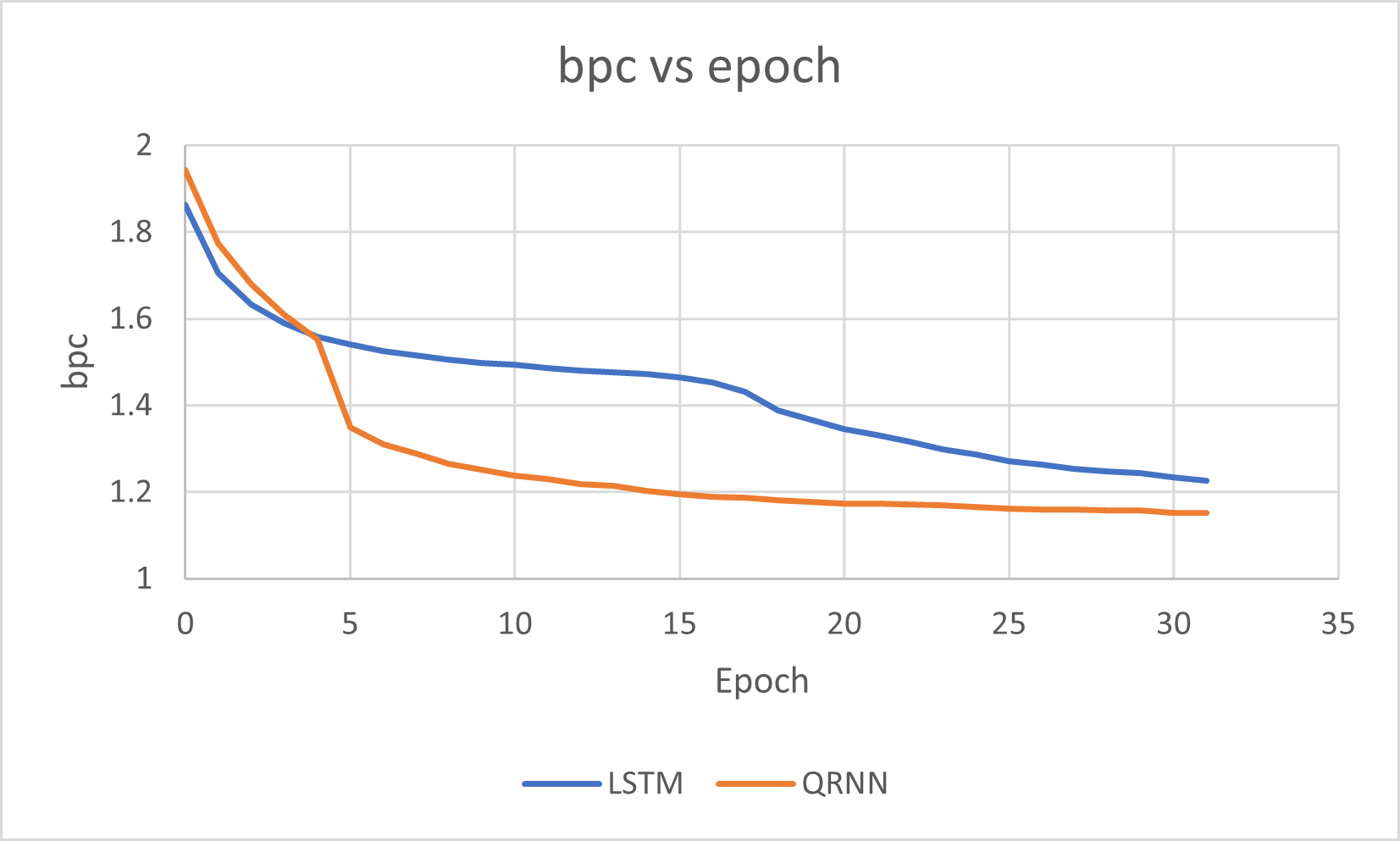}
    \caption{LSTM/QRNN Validation bpc vs epoch on a Tesla V100.}
    \label{fig:lq_bpc_epoch}
\end{figure}

The main advantage of using a QRNN is its speed. QRNNs achieve comparable performance to LSTMs but are much faster. The limitation of LSTMs is their sequential nature, needing to read one token at a time.  The advantage of the QRNN is the parallelizability of the model due to its convolutions. Each convolution can be performed in parallel with different CUDA cores of a GPU which greatly increases the speed of the model~\cite{jagtap_2020}. The lack of parameters to learn in the pooling layer allows for faster sequential calculations as well~\cite{jagtap_2020}. As a result, in our experiments we found there was a noticeable improvement in the training time of the QRNN compared to baseline as shown in figure \ref{fig:lq_bpc_time}.  The QRNN takes less then half the time to train over 32 epochs as compared to the LSTM in the baseline.  Both figure \ref{fig:lq_bpc_time} and figure \ref{fig:lq_bpc_epoch} indicate that there is an improvement in bpc at the end of epoch 31 with the LSTM reaching a value of 1.227 on validation while the QRNN reaches a value of 1.152. 
As stated earlier, the speedup in training time is a result of being able to parallelize the convolutions over the embeddings as well having to learn fewer parameters. Moreover, performing simple pooling calculations as shown in (\ref{ct}) and (\ref{ht}) improve speed. The decrease in bpc is likely a result of the convolutions focusing on more important aspects of the character embeddings which helps in predicting the next byte.  Table \ref{tab:exp_results} shows that the QRNN achieved a bpc of 1.126 on the test set which is much lower than the value of 1.208 achieved by the LSTM.

\begin{table}[]
    \centering
    \resizebox{\columnwidth}{!}{%
    \begin{tabular}{|c|c|c|c|c|}
        \hline \textbf{Experiment} & \textbf{Time/Epoch} & \textbf{Params} & \textbf{loss} & \textbf{bpc} \\
        \hline 4 layer (1) & 1.40h & 54M & 3.52 & 5.07 \\
        \hline 4 layer (2)  & 1.41h & 54M &3.52 & 5.07 \\
        \hline 4 layer (3) base & 1.51h & 54M &0.79 & 1.146 \\
        \hline 4 Layer (4) & 1.41h & 54M &0.79 & 1.141 \\
        \hline 3 Layer (3,3) & 1.15h & 41M &0.80 & 1.16 \\
        \hline 2 Layer (2,2) & 0.84h & 29M &0.81 & 1.164 \\
        \hline 2 Layer (2) & 0.78h & 29M &0.79 & 1.135 \\
        \hline
    \end{tabular}
    }
    \caption{Experiments where the location of the attention head was moved to different layers indicated within parentheses. These tests were performed on an RTX 3090 GPU}
    \label{tab:head_layer}
\end{table}

\begin{table}[]
    \centering
    \begin{tabular}{|c|c|c|c|}
        \hline \textbf{Training Time} & \textbf{Params} & \textbf{Test loss} & \textbf{Test bpc} \\
        \hline 18.5h & 26M & 0.82 & 1.180  \\
        \hline
    \end{tabular}
    \caption{SHAQ results on a Tesla V100.}
    \label{tab:shaq_table}
\end{table}

\section{SHAQ} 

\begin{figure}
    \centering
    \includegraphics[scale=0.2]{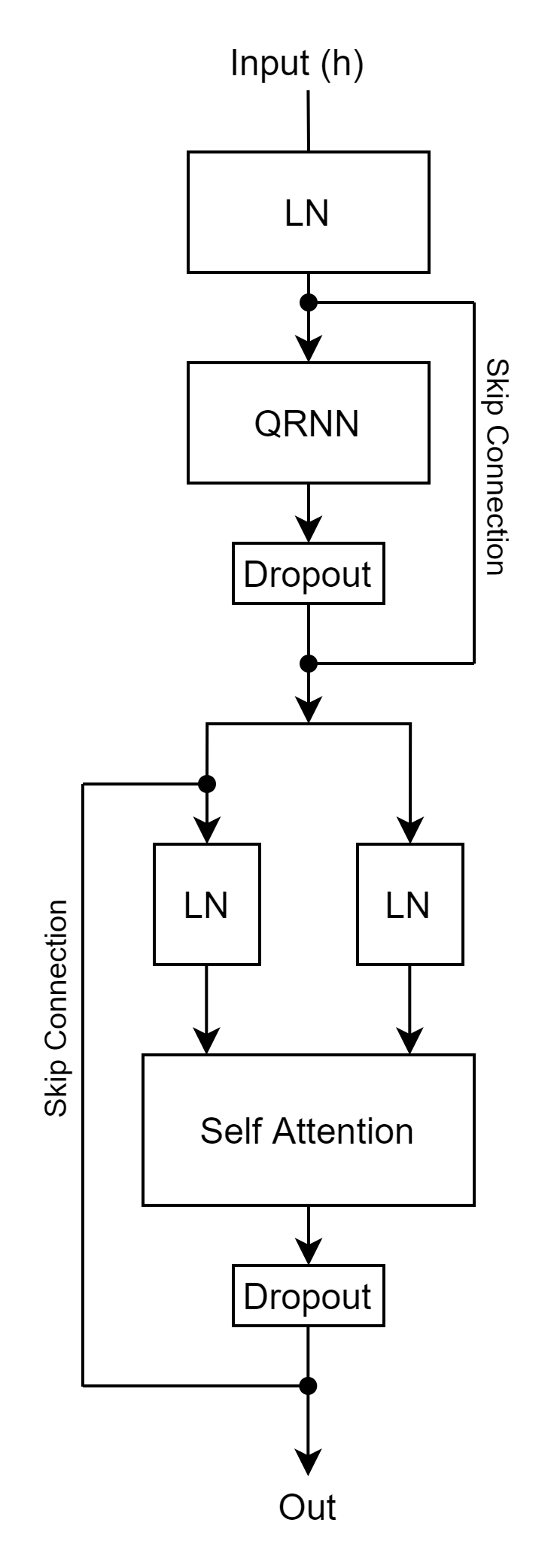}
    \caption{SHAQ Block Diagram}
    \label{fig:shaq}
\end{figure}

Taking everything that was discussed earlier in the paper from our experiments, we decided to combine these findings into a new architecture that we call the single headed attention quasi-recurrent neural network or SHAQ. The basic building blocks are illustrated in figure \ref{fig:shaq}. The figure shows that SHAQ is much simpler compared to SHA-RNN as shown in figure \ref{fig:sha-rnn}. The input is fed into a layer normalization (LN) block which allows for a rescaling of the input to have zero mean and a variance of 1. The output of the LN goes directly into a QRNN cell whose output is then passed through a dropout. The purpose of the dropout layer is to set some of the inputs to 0 to help regularize the network and thus improve the ability of the model to generalize. There is a skip connection on the output of the dropout from the output of the first LN. The purpose of the skip connect is to provide more effective gradients during backpropagation. The result is then fed into two additional LNs and into the simplified self-attention which helps to provide focus on certain characters in the model.  This result is fed through another dropout for further regularization and there is another skip connection to help with gradient flow.  

\begin{figure}
    \centering
    \includegraphics[scale=0.6]{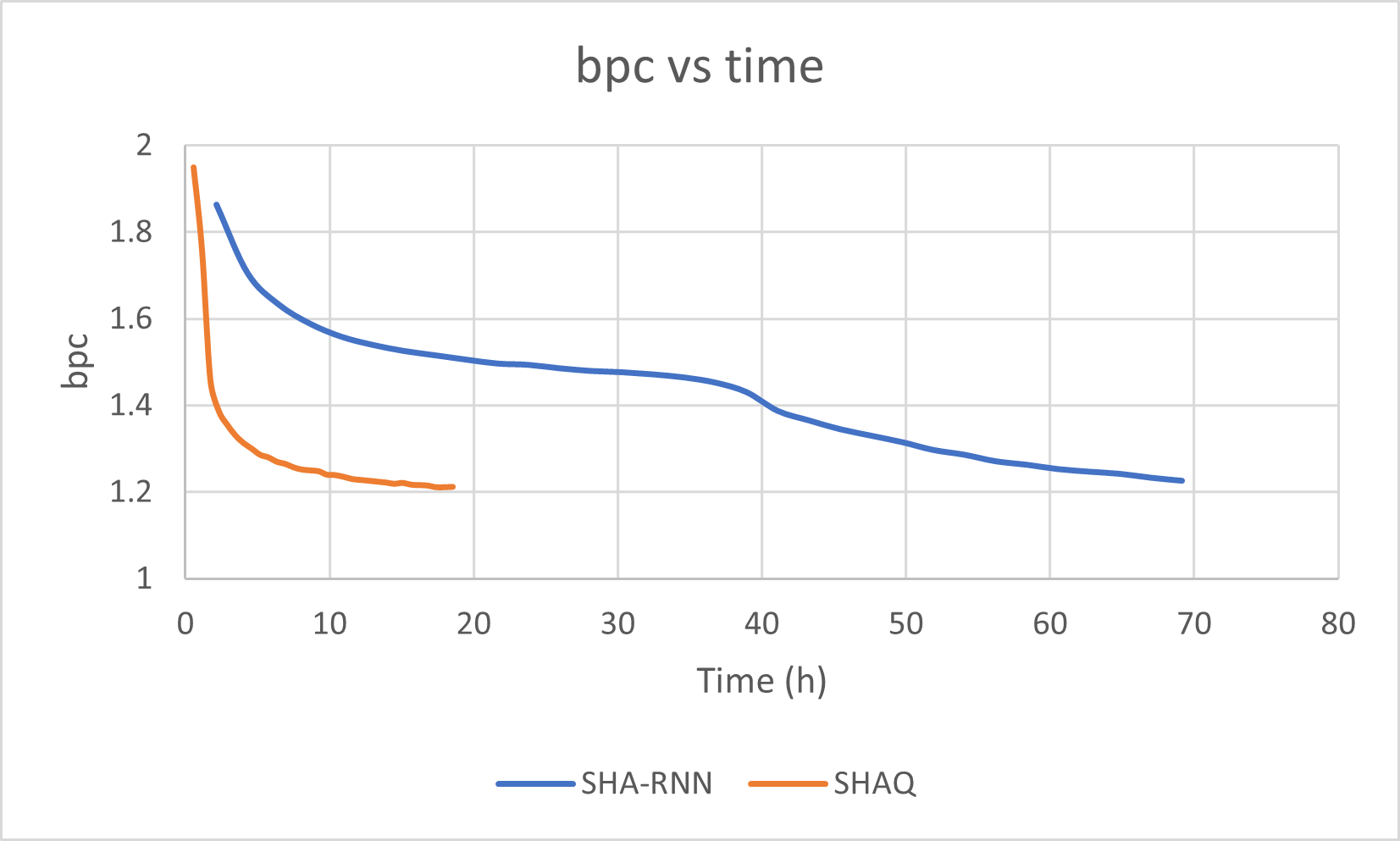}
    \caption{SHA-RNN/SHAQ Validation bpc vs time on a Tesla V100.}
    \label{fig:shaq_bpc_time}
\end{figure}

\begin{figure}
    \centering
    \includegraphics[scale=0.6]{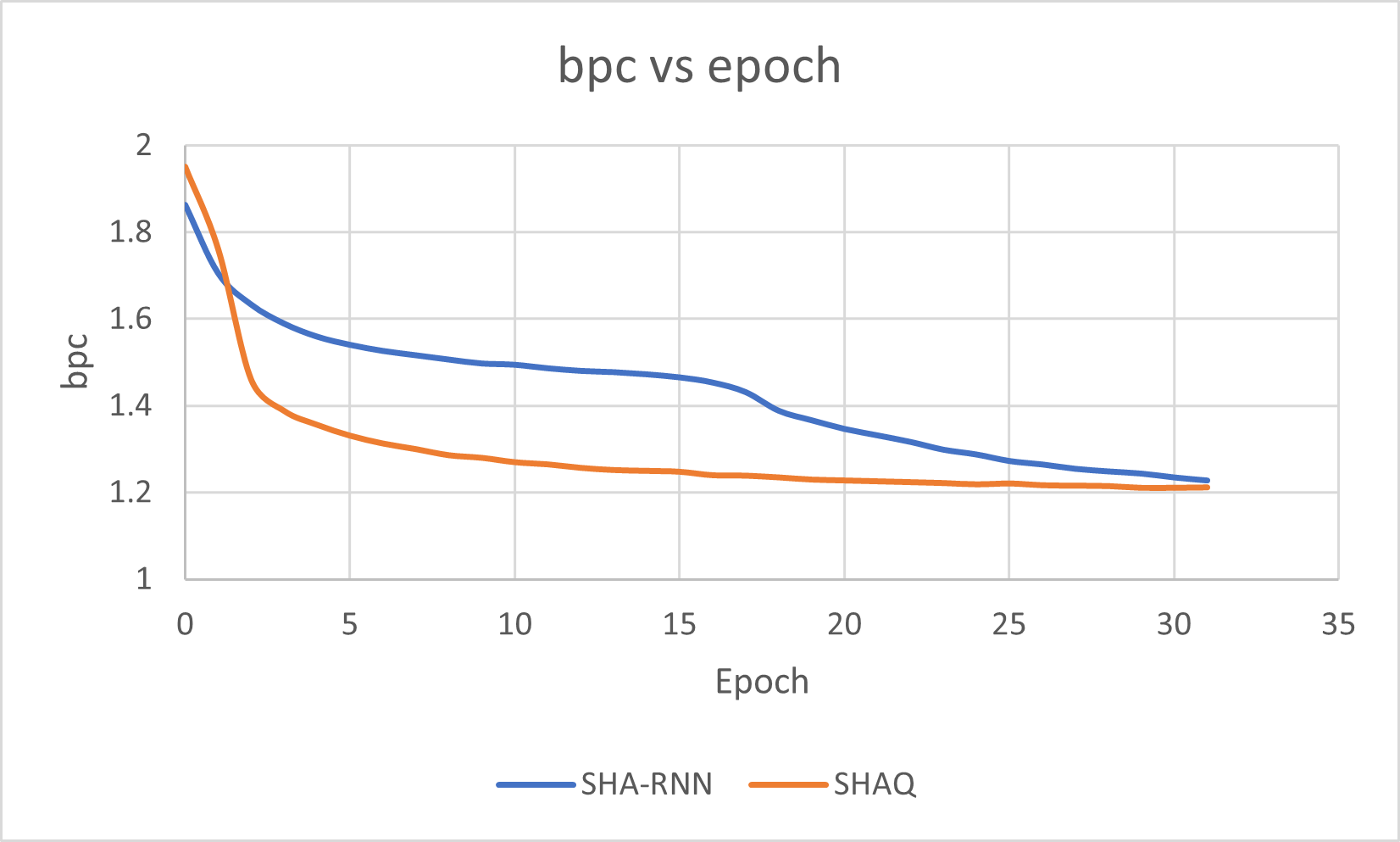}
    \caption{SHA-RNN/SHAQ Validation bpc vs epoch on a Tesla V100.}
    \label{fig:shaq_bpc_epoch}
\end{figure}

Figures \ref{fig:shaq_bpc_time} and \ref{fig:shaq_bpc_epoch} show the results of SHAQ on the validation set of ENWIK8.  Figure \ref{fig:shaq_bpc_time} has SHAQ reaching a final validation bpc of 1.212. This is a good comparison to the baseline validation of 1.208 bpc as shown in figure \ref{fig:lq_bpc_time}. However, SHA-RNN took $\sim$69 hours to train while SHAQ took $\sim$18.5 hours to train, nearly 1/4 of the time. Figure \ref{fig:shaq_bpc_epoch} shows that there is a fairly smooth and steady decline in the bpc value as the number of epochs increase. We were able to reduce the number of parameters by half to 26M from the original 53M of SHA-RNN and SHAQ had 1.18 bpc on test vs 1.208 for SHA-RNN. Table \ref{tab:shaq_table} summarizes the number of parameters of SHAQ and its results on the test set.

\section{Conclusion}

Through our study of SHA-RNN, we discovered that the network still appeared to be too complicated in many aspects and would benefit from simplification. Some aspects of the network could be removed entirely or simplified to result in a less complex network which was able to better generalize over the dataset. In particular, we found replacing the LSTM with a QRNN, removing the Boom layer, removing  attention parameters ($Q_{s}$/$K_{s}$/$V_{s}$), and reducing the depth of the network all improved results over the baseline.

These considerations were combined into a new network called SHAQ, which was able to slightly improve upon the results of the SHA-RNN in ~25\% of the training time.  There are still other improvements that could be done to SHAQ over time.  Some examples include doing more robust hyperparameter tuning such as changing the learning rate or number of epochs or trying new variations on the Boom layer. Section 4.3 explored the idea of reducing the number of blocks used in the model. Both SHA-RNN and SHAQ used 4 blocks connected in sequence to train.  An alteration to SHAQ could simply be reducing the number of blocks while increasing the number of QRNN layers in each block. Since attention is all you need these days, another area to explore is increasing the number of heads per block as well. 


{\small
\bibliographystyle{ieee_fullname}

}

\end{document}